\documentclass{article}

\usepackage[preprint]{neurips_2025}

\usepackage{amsmath}
\usepackage{amssymb}
\usepackage{amsfonts}       
\usepackage{bm}
\usepackage{todonotes}

\usepackage[utf8]{inputenc} 
\usepackage[T1]{fontenc}    
\usepackage{hyperref}       
\usepackage{url}            
\usepackage{booktabs}       
\usepackage{nicefrac}       
\usepackage{microtype}      

\usepackage[nolist]{acronym}
\begin{acronym}[DRAM]
    \acro{ANN}{Artificial Neural Network}
    \acrodefplural{ANN}{Artificial Neural Networks}
    \acro{DNN}{Deep neural network}
    \acrodefplural{DNN}{Deep neural networks}
    \acro{NN}{neural network}
    \acro{ML}{machine learning}
    \acro{MLP}{Multilayer Perceptron}
    \acro{TA}{Tsetlin automaton}
    \acrodefplural{TA}{Tsetlin automata}
    \acro{TM}{Tsetlin machine}
    \acrodefplural{TM}{Tsetlin machines}
\end{acronym}

\title{Uncertainty Quantification in the Tsetlin Machine}

\author{
  Runar Helin \\
  Department of IKT\\
  University of Agder\\
  Grimstad, Norway \\
  \texttt{runar.helin@uia.no}\\
  \And 
  Ole-Christoffer Granmo\\
  Department of IKT\\
  University of Agder\\
  Grimstad, Norway \\
  \texttt{ole.granmo@uia.no}\\
  \And
  Mayur Kishor Shende\\
  Department of IKT\\
  University of Agder\\
  Grimstad, Norway \\
  \texttt{mayurks@uia.no}\\
  \And
  Lei Jiao\\
  Department of IKT\\
  University of Agder\\
  Grimstad, Norway \\
  \texttt{lei.jiao@uia.no}\\
  \And
  Vladimir I. Zadorozhny\\
  School of Computing and Information\\
  University of Pittsburgh\\
  Pittsburgh, PA, USA \\
  \And
  Kunal Ganesh Dumbre \\
  Department of IKT\\
  University of Agder\\
  Grimstad, Norway \\
  \texttt{kunalgd@uia.no}\\
  \And
  Rishad Shafik\\
  School of Engineering\\
  Newcastle University\\
  Newcastle upon Tyne, UK\\
  \texttt{rishad.shafik@newcastle.ac.uk}\\
  \And
  Alex Yakovlev\\
  School of Engineering\\
  Newcastle University\\
  Newcastle upon Tyne, UK\\
  \texttt{alex.yakovlev@newcastle.ac.uk}\\
}

\begin{document}

\maketitle
\begin{abstract}
Data modeling using \acp{TM} is all about building logical rules from the data features. The decisions of the model are based on a combination of these logical rules. Hence, the model is fully transparent and it is possible to get explanations of its predictions. In this paper, we present a probability score for \ac{TM} predictions and develop new techniques for uncertainty quantification to increase the explainability further. The probability score is an inherent property of any \ac{TM} variant and is derived through an analysis of the \ac{TM} learning dynamics. Simulated data is used to show a clear connection between the learned \ac{TM} probability scores and the underlying probabilities of the data. A visualization of the probability scores also reveals that the \ac{TM} is less confident in its predictions outside the training data domain, which contrasts the typical extrapolation phenomenon found in \acp{ANN}. The paper concludes with an application of the uncertainty quantification techniques on an image classification task using the CIFAR-10 dataset, where they provide new insights and suggest possible improvements to current \ac{TM} image classification models.
\end{abstract}

\section{Introduction}
In modern times, many decisions are made based on data models and predictions from a machine learning system. These machine learning models are useful to discover complex patterns and relations in the data. However, complex models, especially those based on \acp{DNN}, are non-transparent. An interesting alternative to \ac{DNN} models is the \ac{TM} \citep{granmo2018tsetlin}, which is a model based on propositional logic and the learning of logical rules from the data features. Since predictions made by the \ac{TM} are combinations of these rules, the model is completely transparent, and hence, interpretable by humans. The \ac{TM} framework supports a wide range of applications, including classification and regression tasks \citep{abeyrathna2019, drop-clause, parallel-tm}, as well as image analysis and signal processing \citep{granmo2019convolutional, jeeru2025interpretable}. It has also been applied in federated learning \citep{qi2025fedtmos}, the contextual bandit problem \citep{bandit}, and natural language processing (NLP) \citep{saha2022relational, rbe-rohan, human-sentiment, tm-ae}. Moreover, \ac{TM} is known for being hardware-friendly due to its energy efficiency \citep{tunheim2024tsetlin, tm-edge}. Its convergence properties have been analyzed in studies such as \citep{jiao2021convergence, zhang2020convergence}.

Despite the many successful applications of \acp{TM}, a good probabilistic understanding is still missing. This paper aims to address that by demonstrating techniques of uncertainty quantification using a novel probability score derived through an analysis of the \ac{TM} learning dynamics. The proposed probability score works as a metric to evaluate certainty in model predictions and opens up new interesting applications. As will be shown in this paper, the uncertainty quantification framework can significantly increase the explainability of the \ac{TM} model, which ultimately leads to a higher trust in its predictions.

The paper is organized as follows. Section \ref{theory} gives an overview of the \ac{TM} and the learning mechanisms relevant to the uncertainty quantification and derives the probability score. Section \ref{results} provides examples of uncertainty estimation using simple learning tasks from simulated data to highlight the connection between the class sums and estimation uncertainty, as shown in Section \ref{simulated_data}. While the experiment in Section~\ref{image_classification} presents the uncertainty quantification applied on an image classification task using the CIFAR-10 dataset \citep{krizhevsky2009learning}. Finally, Section \ref{discussion} concludes with the a discussion of the probability score and ideas for other applications.

\section{Theory}\label{theory}
This section gives a brief overview of how the \ac{TM} works. A complete description of the Tsetlin machine is found in \citet{granmo2018tsetlin}. The \ac{TM} solves a data modeling task by learning patterns in the data. These patterns are represented as sets of conjunctive clauses, voting in favor for or against different targets. The \ac{TM} learns the clauses in a supervised fashion using a specified feedback mechanism described in more details below. Each clause in the \ac{TM} is constructed by a series of \acp{TA}. For each binary feature in the data, there are two \acp{TA}. One for the feature itself and one for its negation. The state values in the \acp{TA} are updated during learning. These values determine whether the literals are included in the clauses.

\subsection{Tsetlin machine learning}
Denote the set of clauses as $\mathcal{C}$. We here assume without loss of generality that the learning task is binary classification. The extension to multiclass problems involves a one-vs-rest approach. The set $\mathcal{C}$ is the union of two disjoint subsets $C^{+}$ and $C^{-}$, corresponding to clauses voting for the target (positive polarity) and against the target (negative polarity) respectively. During inference, the class prediction is found by counting the positive and negative clauses matching each sample. For one sample $\bm{x}_i$, if the positive count is larger than negative, the \ac{TM} decides that the sample is more likely to belong to the positive class, which becomes the final prediction of the sample.

The count of the matching positive clauses subtracted by the count of the matching negative clauses is defined as the \emph{class sum}, and is a central metric in the uncertainty quantification of the TM. Mathematically, it is defined as 
\begin{equation}
  v(\bm{x}_i) = \sum_{k=1}^{n_{+}} C^{+}_k(\bm{x}_i) - \sum_{k=1}^{n_{-}} C^{-}_k(\bm{x}_i),
\end{equation}
\noindent where
\begin{equation}
  C_k(\bm{x}_i) = 
    \begin{cases}
      1 & \quad \text{if } \bm{x}_i \text{ matches clause } C_k\\
      0 & \quad \text{otherwise},
    \end{cases}
\end{equation}
\noindent and $n_{+}$, $n_{-}$ are the number of positive and negative clauses respectively. In this paper, we use a modification of the clause counting scheme, where additional variables $w_j \in \mathbb{N}$ called \emph{weights} are included for each clause $j$. The weights are updated during the learning and the class sum is defined as the weighted sum of matching clauses as follows:
\begin{equation}
  v(\bm{x}_i) = \sum_{k=1}^{n_{+}} w^{+}_k C^{+}_k(\bm{x}_i) - \sum_{k=1}^{n_{-}} w^{-}_kC^{-}_k(\bm{x}_i),
\end{equation}

During learning, the \ac{TM} updates the states of each \ac{TA} with a probability that depends on the class sum $v(\bm{x})$ and a hyperparameter $s$ termed the \emph{specificity} in the \ac{TM} literature. Another central parameter is $T$, called the \emph{target value}. This parameter acts as a limit to how large the class sum can get. By adjusting $T$, one encourages the model to learn more diverse clauses. During learning, the class sum is clipped to the range $-T$ and $T$. Therefore, in the rest of this section it is assumed that the class sum is clipped in the same manner.

The clause updates of the \ac{TM} are divided into three different types: recognize, erase and reject. Every sample $\bm{x}_i$ in the training data triggers an update of the TM. If the sample belongs to the target, recognize or erase feedback is given to a clause with probability $P_{I}(\bm{x}_i) = \frac{1}{2}\left(1 - \frac{v(\bm{x}_i) }{T}\right)$. If the sample does not belong to the target class, the clause is given reject feedback with probability $P_{II}(\bm{x}_i) = \frac{1}{2}\left(1 + \frac{ v(\bm{x}_i) }{T}\right)$. Note here that $P_{I}(\bm{x}_i) + P_{II}(\bm{x}_i) = 1$.

\subsection{Uncertainty quantification}
The probability for a clause to receive the two types of feedback depends on whether it is a positive clause $C^{+}$ or a negative clause $C^{-}$. Additionally, the type of feedback is dependent on the value of the true response value $y_i$ associated with the sample $x_i$. Combining these aspects, the complete probability for a positive clause to receive Type I and Type II are

\begin{equation}\label{eq:typeI_prob_complete}
    P(\text{I}|\bm{x}) = P_{I}(\bm{x}_i)P(y=1|\bm{x}),
\end{equation}
\noindent and
\begin{equation}\label{eq:typeII_prob_complete}
    P(\text{II}|\bm{x}) = P_{II}(\bm{x}_i)P(y=0|\bm{x}).
\end{equation}

For a negative polarity clause, the two probabilities are opposite. 

During training, the clauses gets feedback in accordance with the \ac{TM} predictions. Here we describe how the feedback affects the class sum of one particular sample $x$ during training. Broadly speaking, Type I feedback is given to solidify correct predictions while correcting false negative predictions. If this feedback is given to the clauses after seeing the sample, the class sum for that sample tends to increase. On the other hand, Type II feedback is used to correct for false positive predictions and tend to decrease the class sum for the sample. A larger class sums means that the sample produces less Type I feedback and simultaneously produces more Type II feedback. As a result, there is a point where the probabilities of the two feedback types, given a sample $x$, is the same. To find this class sum equilibrium, we set Eq.~(\ref{eq:typeI_prob_complete}) and Eq.~(\ref{eq:typeII_prob_complete}) equal to each other and solve for the class sum $v(x)$ to obtain 
\begin{equation}
    v = T\left[2P(y=1 | \bm{x}) - 1\right].
\end{equation}
From this equation, one find that the probability $P(y=1 | x)$ for the target to be true given a sample can be estimated by the equation: 
\begin{equation}\label{eq:class-uncertainty}
    P(y=1 | x) = \frac{1}{2}\left(1 + \frac{v(\bm{x}_i) }{T}\right),
\end{equation}
\noindent which is coincidentally the same probability as receiving Type II feedback. We call the output of this function the \emph{probability score}. Later, it is shown how this score can be used to quantify the uncertainty of predictions.

\subsection{Alternative derivation of probability score}
Note here that the derivation of Eq.~\ref{eq:class-uncertainty} assumed that the two types of feedback is in a balance. There is an alternative way to derive an equation to convert class sum values to probability scores based on simple scaling and shifting arguments. Since the class sum is clipped at $-T$ and $T$, you first scale the class sum by $T$ to get values in the range -1 and 1. Then, you shift the score to the range 0 and 2 by adding 1. Finally, you divide that by 2 to get the range of 0 and 1 needed for a valid probability score. Putting all these steps together, you get the equation $\frac{1}{2}\left(1 + \frac{v(\bm{x})}{T}\right)$, which is the exact same equation as Eq.~(\ref{eq:class-uncertainty}) which was derived by setting TypeI and TypeII feedback probabilities equal each other. In addition to being the resulting formula by balancing the feedback types, the formula is also the most natural way to convert the class sums to values that can be interpreted as probabilities.

A different choice to convert class sums to probability scores would be the sigmoid activation function (or softmax for multiclass). However, since the derived probability scores is directly connected to the learning dynamics of the \ac{TM} and it is a simple linear transformation, Eq.~(\ref{eq:class-uncertainty}) the most natural choice for the \ac{TM} model.

\section{Results} \label{results}
This section presents examples of applications of the probability score derived in Section \ref{theory}. First, Section \ref{simulated_data} explores properties of the probability scores using simulated data. It begins with the simplest scenario using a dataset of only a single pattern before increases the complexity to a dataset with three features. After that, a dataset of two continuous features is used to demonstrate useful visualizations of uncertainty. Section \ref{image_classification} uses real data to explore uncertainty quantification of a image classification task. All experiments except the image classification were run using a Macbook Pro 2023 edition with 8 CPU cores, 16 GB of RAM and a Apple M3 chip. The image classification was run on an HPC with 1.4 TB of RAM, using a Tesla V100-SXM3 GPU. The training time for the simulated data was all less than one minute and is not reported in more detail, while the training time for the image classification is stated in Section~\ref{image_classification}.

\subsection{Simulated data}\label{simulated_data}
The aim of the first simulated data experiment was to observe how the class sum of a single sample/pattern behaves during training when the data is subjected to different levels of noise. The training data consists of 100 copies of the same sample with three binary features $x=[1,0,1]$. We run in total 11 experiment with varying amount of noise. Here, the noise is the probability of the target being false. For example $P(y=1) = 1.0$ means no noise while $P(y=1) = 0.5$ means 50\% noise. The \ac{TM} parameters were T=2000, s=1.0 and number of clauses set to 20.
\begin{figure}[!htbp]
    \centering
    \includegraphics[width=0.8\linewidth]{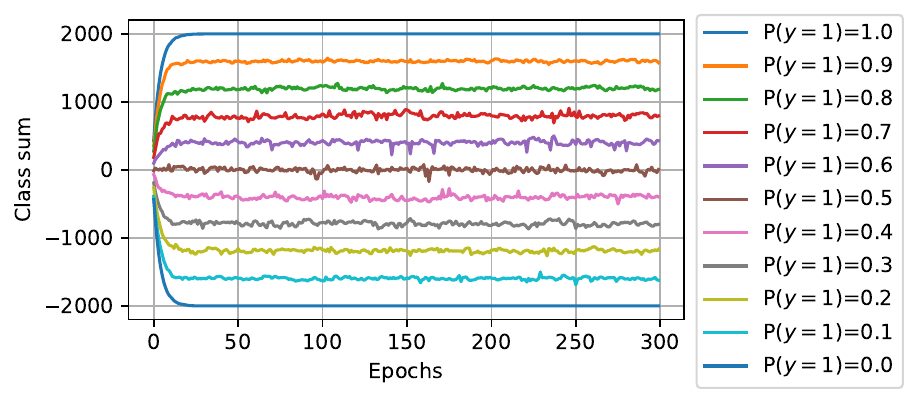}
    \caption{Evolution of class sums for the single pattern learning. }
    \label{fig:single_pattern_learning}
\end{figure}

From the results shown in Figure~\ref{fig:single_pattern_learning}, it is observed that the class sums for each noise level start from zero and, after a few epochs, reach a certain value where they start to oscillate. These converging values are different for each noise level and decrease according to the amount of noise added to the data. When using Eq.~(\ref{eq:class-uncertainty}) on the average of the last 200 epochs, the resulting probability scores are approximately the same as the noise levels (not shown here).

The next simulated data experiment increases the complexity by including samples with different patterns. The data has three binary features, resulting in a total of 8 unique samples. Naming the features $A$, $B$ and $C$ and the target variable $Y$, the underlying data structure is given by the conditional probability table shown in Table~\ref{tab:1}.
\begin{table}[htbp]
  \centering
  \caption{Conditional probability table for the eight samples with three features.}
  \label{tab:1}
  \begin{tabular}{lllllllll}
    \toprule
    $A$      & 1 & 1 & 1 & 1 & 0 & 0 & 0 & 0\\ 
    $B$      & 1 & 1 & 0 & 0 & 1 & 1 & 0 & 0\\ 
    $C$      & 1 & 0 & 1 & 0 & 1 & 0 & 1 & 0\\
    \midrule
    $P(Y=1)$ & 1.00 & 0.90 & 0.80 & 0.60 & 0.40 & 0.20 & 0.10 & 0.00\\
    \bottomrule
  \end{tabular}
\end{table}

\begin{figure}[htbp]
    \centering
    \includegraphics[width=\textwidth]{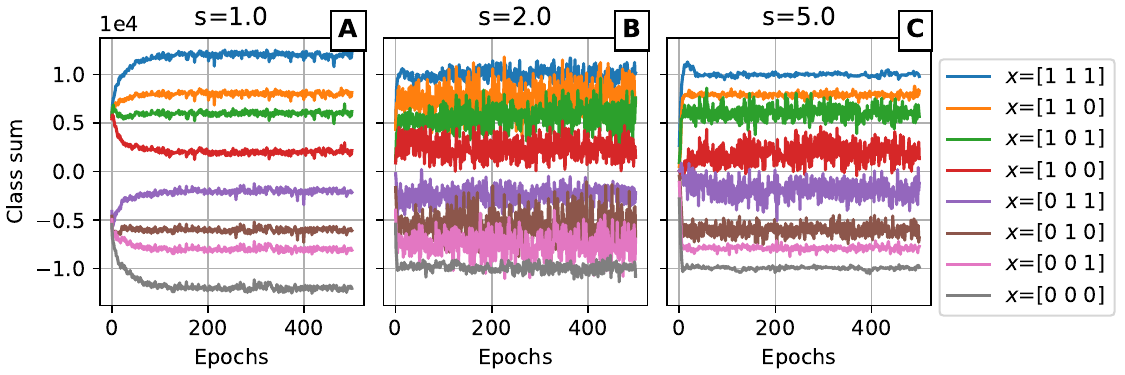}
    \caption{Evolution of class sums for the experiment with eight patterns and different $s$ values.}
    \label{fig:eight_pattern_learning}
\end{figure}

Different from the single-pattern experiment each sample is here associated with different amounts of noise. Figure~\ref{fig:eight_pattern_learning} shows the results where a similar trend of converging class sums is happening. This is further evidence for the connection between the class sums and the underlying probabilities. A high class sum is clearly correlated with the amount of noise, which can be interpreted to mean that higher class sums corresponds to more certain predictions. This experiment also shows that a high class sums for a samples is only possible whenever there is small amount of noise present.

This experiment also includes results using different values of the $s$-parameter in the \ac{TM} model. The choice of $s$ is affecting the class sums both in terms of oscillation strength and average value. A thorough analysis of this effect is outside the scope of this paper, but it is most likely related to the amount of shared features of the clauses in the models with different $s$. As explained in \citet{granmo2018tsetlin}, a smaller $s$ means that the clauses include fewer literals. Therefore, it is hypothesized, the clauses of the model with $s=1.0$ is more general and is shared among more samples compared with the other models. Therefore, the class sums of each sample is affected by feedback from the other samples differently depending on the $s$ value. This is supported by the fact that we found the number of unique clauses of the models to be 6, 15 and 22 for $s$ values equal to 1.0, 2.0 and 5.0 respectively. Despite the differences caused by the $s$ values, the order of the class sums of the different samples is the same, meaning that the interpretation of the probability score calculated from the class sums as uncertainty is still valid.

\subsubsection{Visual assessment of uncertainty}\label{uncertainty_region}
\begin{figure}[htbp]
  \centering
    \includegraphics[width=0.7\textwidth]{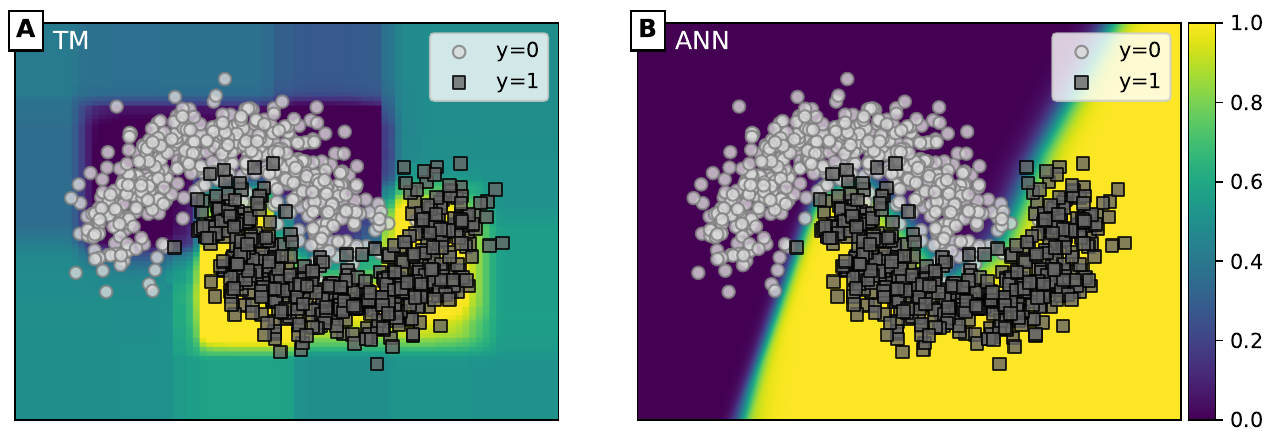}
    \caption{Prediction certainty by region of \ac{TM} (a) and \ac{ANN} (b) models.}
     \label{fig:uncertainty_regions}
\end{figure}

In this experiment we used simulated data of two continuous features from the \texttt{make\_moons()} function in scikit-learn~\citep{scikit-learn}. When plotted together, the data forms two half-moons where the objective is to classify the samples to the correct one. The two half-moons are overlapping by a small amount as seen in the distribution in Figure~\ref{fig:uncertainty_regions}, so a perfect separation is not possible. The dataset was balanced and had 1000 samples in total. The \ac{TM} model in this experiment used parameters $T=10,000$, $s=1.1$ and 1000 number of clauses. The model was trained for 15 epochs.

To make this data applicable for the \ac{TM} models, the features have to be binarized. This was done by dividing each feature into 64 uniform bins and apply a thermometer encoding. More specifically, each bin is represented by a value in a bit-string. A bit with value of 1 in the encoded bit-string means that the feature value is greater or equal the corresponding feature threshold value of that bin. One additional bit per feature was added at the end of the encoded bit-strings corresponding to the largest value of that feature in the training data. This extra bit allowed for encoding of feature values outside the training data domain. More specifically, feature values smaller than the minimum value will be bit-strings of all zeros while feature values larger than the maximum value will bit-strings of all ones.

To visualize uncertainty regions we created a mesh grid of values covering the feature space and the surrounding area. These data points were then transformed into binarized features and predicted by the trained \ac{TM} model. Using Eq.~(\ref{eq:class-uncertainty}), the class sums were converted to probabilities and plotted together with the data as shown in Figure~\ref{fig:uncertainty_regions}. The figure also includes a similar plot for a \ac{MLP}, where the probability score is the output of the output layer using a \textit{sigmoid} activation function. The \ac{MLP} model here has one hidden layer with 32 nodes and rectified liner unit (ReLU) activation. At each point in the mesh grid the estimated probability score for class $y=1$ is plotted with the score value indicated by the background color. The data distribution is also shown as a reference of the training data. For the \ac{TM} model in Figure~\ref{fig:uncertainty_regions}, the regions in the vicinity of the data points have probability scores close to 0 or 1, indicating a certainty in those predictions as discussed above. At the border between the two half-moons, the probability is closer to 0.5 (i.e., green color), indicating a high level of uncertainty. Additionally, the regions outside the domain of the training data also have probability scores close to 0.5. This shows that the \ac{TM} is not extrapolating outside the training data, which is in contrast to a typical \ac{ANN} model as shown in Figure~\ref{fig:uncertainty_regions}b.
\begin{figure}[htbp]
     \centering
     \includegraphics[width=0.7\textwidth]{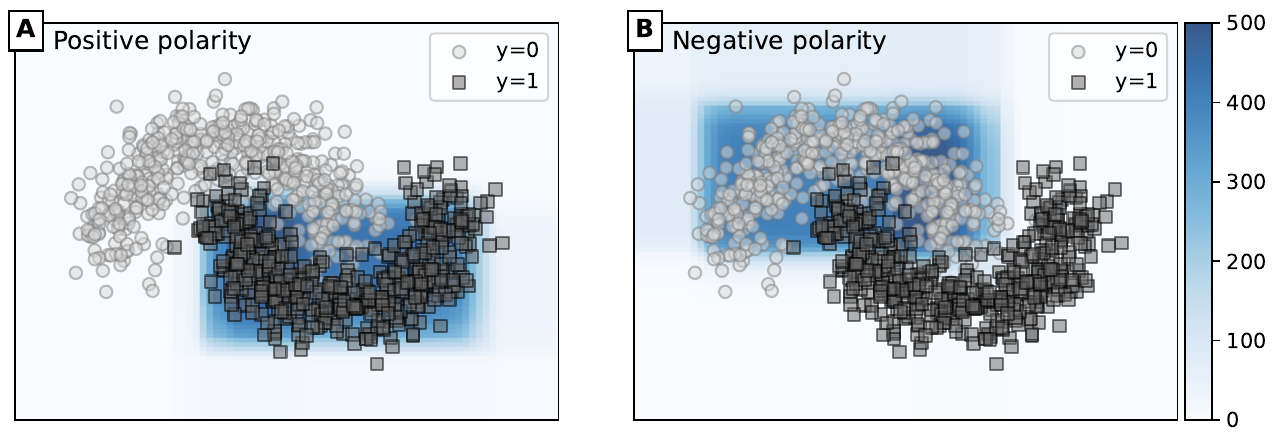}
     \caption{Visualization of clause counts by region for positive polarity (a) and negative polarity (b) clauses.}
     \label{fig:clause_counts}
\end{figure}
A possible explanation of why the \ac{TM} model is not extrapolating to the same degree is related to the fact that the literals corresponding to the regions outside the training domain appear very infrequently in the \ac{TM} clauses, because there is little useful information there regarding the classification task. It should be noted that when using a larger $s$ value the probability scores outside the training domain tended to increase. The reason for this is that the clauses include more literals when $s$ is larger, as explained in the previous section, meaning that the clauses tend to include the literals for the boundary thresholds more frequently. This can happen because it will not affect the prediction accuracy since those literals are constant for all training samples. Therefore, it is desired with a small $s$ values for applications similar to this experiment.

\subsection{Uncertainty quantification in image classification}\label{image_classification}
This section explores how the class sum analysis can be used on a image classification task using the CIFAR-10 dataset \citep{krizhevsky2009learning}. This is a challenging task for the \ac{TM} model with the current best-performing \ac{TM} model reaching a 82.8\% test set accuracy using a form of an ensemble model called the composite \ac{TM} \citep{gronningsaeter2024}. For a single convolutional \ac{TM} model, the expected test set accuracy is in the 60-70\% range, depending on the size of the model.

\subsubsection{Dataset and training}
The data consists of 50,000 training samples and 10,000 test samples of $32 \times 32 $ sized color images from 10 different classes. Before training the \ac{TM} model, all the images were binarized using a thermometer encoding with resolution 8 for each color channel similar to \citet{gronningsaeter2024}. The main results in this paper used a convolutional version of the \ac{TM} with 2000 clauses per class, $T=20,000$, $s=20$ and a patch size of $3\times3$ pixels. Additionally, a literal budget of 64 was used to obtain shorter clauses \citep{abeyrathna2023}. The model was trained for 100 epochs and achieved a test set accuracy of 65\%. An additional experiment was conducted with the exact same parameters except the parameter $s=1.2$. The smaller $s$-value resulted in lower accuracy of 63\%. For each model, the training took 5 minutes per epoch for a total training time of around 8 hours for all 100 epochs on a single GPU.

\subsubsection{Certainty quantification}
\begin{figure}[htbp]
    \centering
    \includegraphics[width=0.8\linewidth]{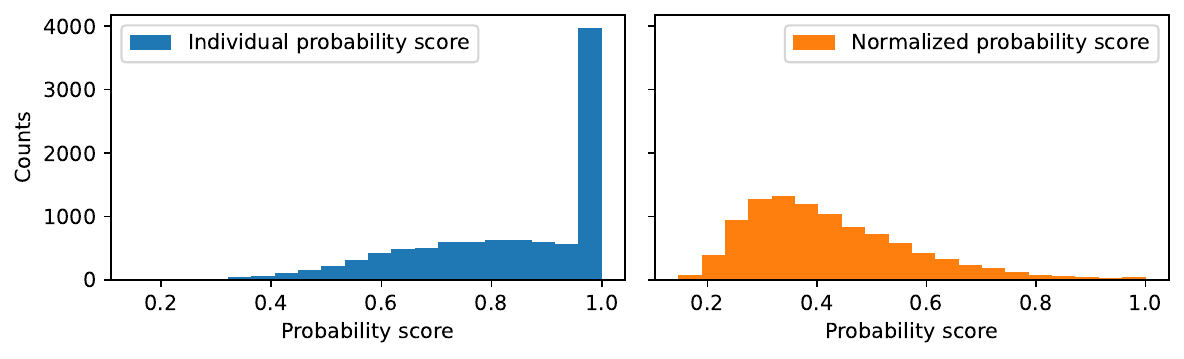}
    \caption{Histograms of the individual and normalized probability scores of the CIFAR-10 test set samples.}
    \label{fig:certainty_hist}
\end{figure}

When considering the probability scores for the CIFAR-10 problem, it became apparent that the score needed a modification to handle multiclass problems. The reason is that the multiclass-\ac{TM} treats each class as a binary classification problem, each with its own class sum, and where the prediction is the class with the highest class sum. If the probability score was applied on just the class with the highest class sum, the score would reflect the certainty of a binary one-vs-rest classification and not consider the certainty of the other classes. To address this, we propose a normalization of the probability scores across all classes as a modification. More specifically, the normalized probability score is the largest individual probability score divided by the sum of the probability scores for all classes. The rationale behind this choice is that a prediction is more certain if it has high certainty for the predicted class (implying a high class sum) while simultaneously has low certainty for all the other classes. A high normalized probability score therefore means that the \ac{TM} has found patterns that clearly separates the predicted class from any other class. Figure~\ref{fig:certainty_hist} illustrates the difference between the individual and normalized class sums.
The figure shows the distributions of the two probability scores for the test set samples. If the uncertainty is judged from the individual probability scores, the predictions would be attributed a high certainty. This interpretation seems wrong in light of the low accuracy of only 65\%. A more accurate picture is obtained from the normalized probability scores, which is more uncertain about the majority of the samples. Therefore, all probability scores for this application are the normalized scores in the rest of the paper.

\begin{figure}[ht]
    \centering
    \includegraphics[width=0.7\linewidth]{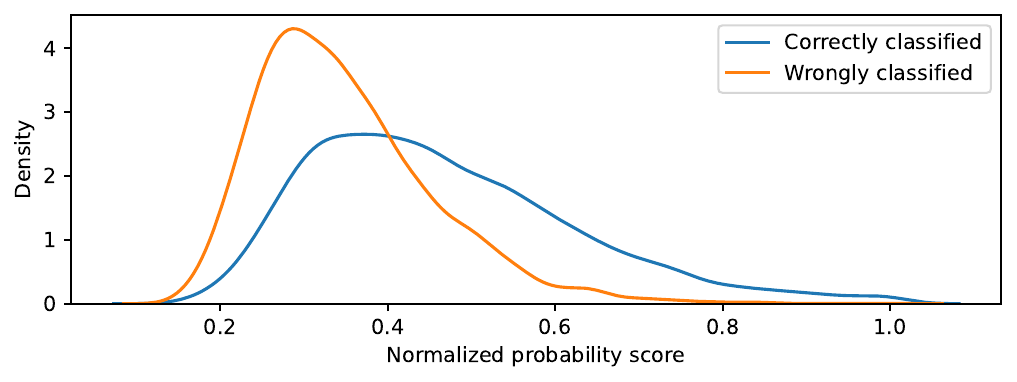}
    \caption{Kernel density estimation of max class sum distributions of CIFAR-10 test set predictions.}
    \label{fig:cifar-kde}
\end{figure}

To assess the validity of the probability score on this dataset we plot the distributions of the normalized probability scores for correctly and incorrectly samples as shown in Figure~\ref{fig:cifar-kde}. Since the correctly classified samples have a higher expected probability score, this plot verifies that a higher certainty yields more accurate predictions.

\begin{figure}[ht]
    \centering
    \includegraphics[width=0.8\linewidth]{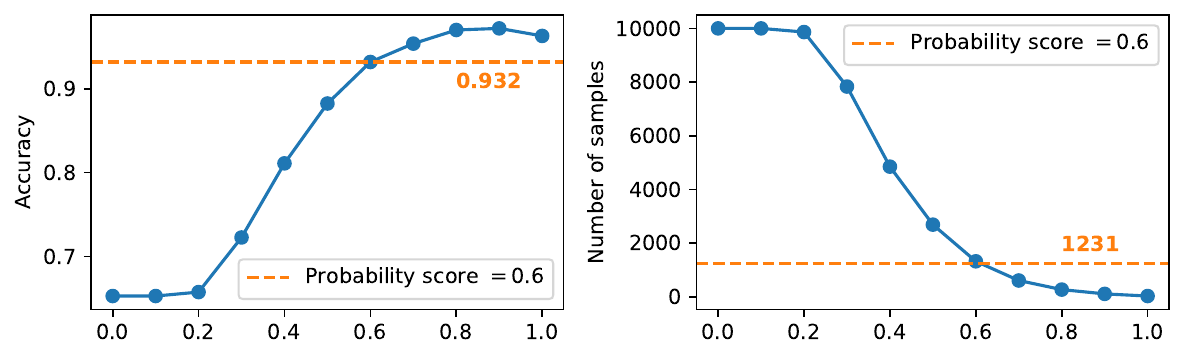}
    \caption{Accuracy and number of samples of subsets of test set samples with different levels of certainty.}
    \label{fig:cifar-max-cs}
\end{figure}

To further explore the certainty of predictions we calculate the accuracy of subsets of samples with with different degree of certainty. The results are shown in Figure~\ref{fig:cifar-max-cs}.
Each point in the figure shows the the accuracy calculated from the subsets of all the samples with greater or equal normalized probability score than each threshold level, ranging from 0.0 to 1.0. Here it is clear that when considering only the samples with a high probability score, the accuracy increases. For example, if one take all samples with probability score greater or equal to 0.6, the accuracy is 0.932. The number of samples with this degree of certainty is 1,231 of a total of 10,000 samples. This suggests an interesting application of the probability score when one require high certainty in the model. One can use the probability score as a threshold and only accept predictions with a greater certainty than a desired threshold. It also reflects how good the model is performing by looking at the number of samples with the required certainty. If that number is too small, it signifies a poor model and could be a sign that more data is needed or you need a different set of model parameters.

\subsubsection{Visualization of samples}
\begin{figure}[h]
    \centering
    \includegraphics[width=0.7\linewidth]{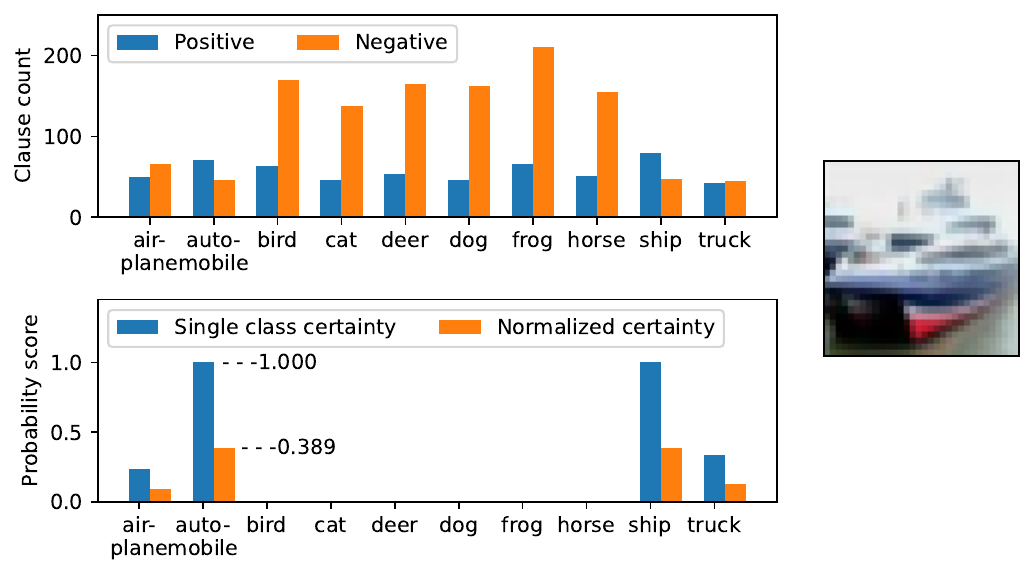}
    \caption{Clause counts of per class of a sample in the test set with label \textit{ship}.}
    \label{fig:cifar10_sample}
\end{figure}

More insights are gained if one look at the number of active clauses, similar to what was done in Section \ref{uncertainty_region}. Figure~\ref{fig:cifar10_sample} displays the clause count and probability scores for a single sample to demonstrate how this information can be used.
One observation is that all classes have activated negative polarity clauses for all classes, even the correct class. The animal classes have the most activated negative clauses, meaning that the \ac{TM} model finds strong evidence for this sample not being an animal. On the other hand, it finds less evidence voting for any specific class. Ship is the class with the most active positive clauses followed by automobile and frog. Since the majority of the active clauses for classes automobile and ship is positive, the class sum for these classes is the largest. In fact, judged individually, both the class sums are above the threshold value $T$. The normalized probability score of 0.389 on the other hand reflects the fact that the model is not sure which of the two classes that is correct.

\begin{figure}[htbp]
    \centering
    \includegraphics[width=\linewidth]{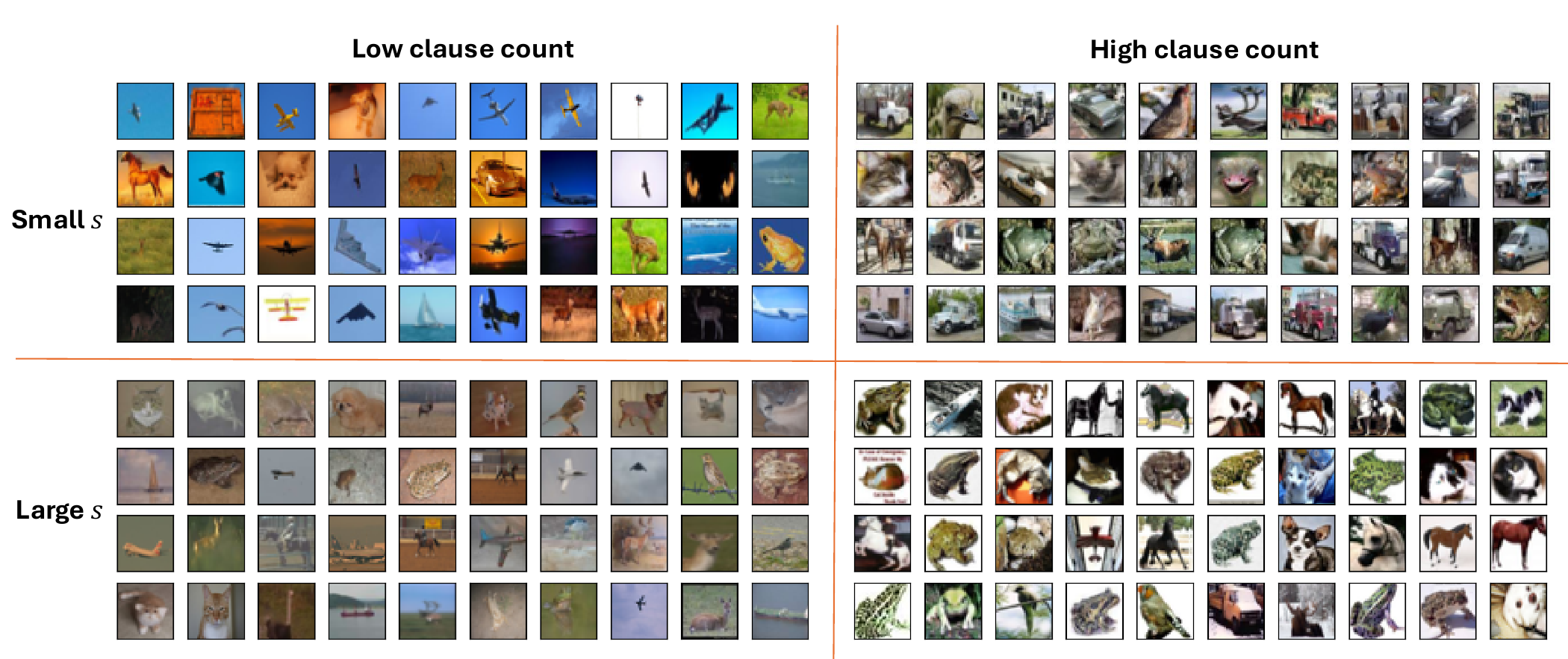}
    \caption{Illustration on CIFAR-10 test images with different number of active clauses.}
    \label{fig:clause_count}
\end{figure}

The clause count provides information about the number of patterns the \ac{TM} model has learned that match the sample. If the count is high, it means that the sample contains many common features that either recognize the class or discriminate a class. Samples with low clause counts do not have many of these common features and are therefore expected to be notably different from the samples with high clause counts. Figure~\ref{fig:clause_count} shows groups of samples with either low or high clause counts and for two different \ac{TM} models where only the $s$-parameter is different.
Each group is the 40 samples in the test set with the lowest/highest clause count for each model. Focusing on the model with large $s$, the samples with a low clause count are more diffuse and have a smaller degree of contrast within each image compared with the group with a high clause count. In the model with small $s$ there is also a difference. Here, the samples with low clause count are more monochromatic and have fewer details. A possible explanation for the difference is related to the binarization using the thermometer encoding. Samples with lower contrast have most pixel-values concentrated in the same pixel-value range, making the information end up in fewer bins in the thermometer encoding compared to the high-contrast images. In other words, there is less information to use and learn for the \ac{TM} model in the low-contrast images. The same argument holds for the more monochromatic samples. To address these potential weaknesses, a form for preprocessing such as histogram normalization could improve the model by making samples more similar. This is an interesting new direction for future study to improve the image classifications of \ac{TM} models.

\section{Discussion}\label{discussion}
This paper have used examples to demonstrate the connection between the probability score and model uncertainty, both using simulated and real data. This is a first step towards uncertainty quantification of \ac{TM} models. Naturally, there are some current limitations that will is briefly discussed here.

\textbf{Limitations.} The presented probability score for a single sample does not truly converge to a single value but oscillates and changes from epoch to epoch, as seen in the plot in Figure~\ref{fig:eight_pattern_learning}. This is an effect of the continuous learning property of the \ac{TM} model. It means that at any epoch, the probability score could be smaller or larger than the expected score. A possible way to get a more accurate probability score is therefore to collect certainty scores for the test set for several epochs while training the model. This can safely be done without information leakage since the calculation of the probability scores does not require the true label. The process of averaging certainty score would therefore makes inference slower since you have to let the \ac{TM} train while collecting probability scores, but it could be a worthwhile trade-of if more accurate certainty metrics are needed.

The paper also does not address uncertainty in regression \ac{TM} models. It is not obvious how to modify this approach for this and is something that will be explored in future research.

\textbf{Future works}
The connection between probability score and uncertainty could be interesting to apply in the context of causal discoveries and Bayesian networks. Another possibility is to use the probability scores in a framework for a generative \ac{TM} model. An extension of uncertainty quantification for regression task will also be a an interesting subject for future research.

\section{Conclusions}
In this work we have presented a probability score that is applicable for any \ac{TM} model which can be used as an additional metric during inference. It gives you information about the certainty of the \ac{TM} model and could help in deciding how much confidence you should have in the prediction. The score can also work as a threshold for a required certainty level, useful for example in healthcare where transparency and accuracy of predictions are essential. In combination with the clause count, the probability score is a powerful tool for uncertainty quantification of \ac{TM} models.

\bibliography{references}
\bibliographystyle{plainnat}

\end{document}